# Broad Critic Deep Actor Reinforcement Learning for Continuous Control

Shiron Thalagala, Pak Kin Wong, Xiaozheng Wang, and Tianang Sun

*Abstract*—**In the domain of continuous control, deep reinforcement learning (DRL) demonstrates promising results. However, the dependence of DRL on deep neural networks (DNNs) results in the demand for extensive data and increased computational cost. To address this issue, a novel hybrid actor-critic reinforcement learning (RL) framework is introduced. The proposed framework integrates the broad learning system (BLS) with DNN, aiming to merge the strengths of both distinct architectural paradigms. Specifically, the critic network employs BLS for rapid value estimation via ridge regression, while the actor network retains the DNN structure to optimize policy gradients. This hybrid design is generalizable and can enhance existing actor-critic algorithms. To demonstrate its versatility, the proposed framework is integrated into three widely used actor-critic algorithms– deep deterministic policy gradient (DDPG), soft actor-critic (SAC), and twin delayed DDPG (TD3), resulting in BLS-augmented variants. Experimental results reveal that all BLS-enhanced versions surpass their original counterparts in terms of training efficiency and accuracy. These improvements highlight the suitability of the proposed framework for real-time control scenarios, where computational efficiency and rapid adaptation are critical.**

*Index Terms*—**Broad critic deep actor, broad learning system, broad reinforcement learning, continuous control, deep reinforcement learning.**

## I. INTRODUCTION

THE application of reinforcement learning (RL) in continuous control is promising especially in solving complex engineering tasks [1]. Some engineering control problems solved by RL are robotics control [2], autonomous driving [3], industrial process control [4], and aeronautics [5]. In such applications, assessing the effectiveness of RL agents requires crucial consideration of both learning efficiency and computational cost, particularly in the context of online training [6].

Recent research suggests deep reinforcement learning (DRL) for successful online continuous control [7]. In DRL, deep neural networks (DNNs) are utilized for approximating both policy and value functions. However, DRL introduces certain challenges attributable to the inherent properties of DNNs [8]. Significantly, these networks require high volumes of data samples to effectively train and refine policies. This requirement becomes especially evident in situations involving complex environments with high-dimensional state-action spaces. Gathering and processing extensive data in real-time in such scenarios can lead to significant computational cost and extended training period. Moreover, to achieve higher accuracy, additional layers need to be added to DNNs in DRL algorithms. This requires re-estimating parameters in all existing layers through gradient descent algorithms which leads to time-consuming model updates. These problems can greatly affect the training efficiency of complex continuous control problems.

To address these challenges, one promising approach is to explore the opportunity of applying broad learning systems (BLS) [9] to RL. BLS networks are characterized by relatively shallow architecture in comparison to the deep architecture in conventional DNNs. Compared to DNNs, the BLS shows different advantages such as less complexity of the structure and less computational time during training. Moreover, the incremental learning ability in BLS eliminates the requirement of remodeling the network in order to enhance accuracy.

In this study, a hybrid approach is proposed where both deep and broad networks are utilized in actor-critic RL [10], which is referred to the broad critic deep actor (BCDA) framework. The primary idea behind this framework is to replace the existing DNN-based critic with BLS in actor-critic algorithms designed for continuous control tasks, such as deep deterministic policy gradient (DDPG) [11], soft actor-critic (SAC) [12], and twin delayed deep deterministic policy gradient (TD3) [13], in order to improve learning efficiency while maintaining accuracy. Additionally, when the BCDA framework is applied to a certain algorithm, the critic network parameter estimation uses ridge regression, while the actor network parameters are updated using the original method employed by each algorithm.

At the same time, recent research has demonstrated the effectiveness of applying BLS in DRL, where BLS networks replace DNNs in DRL algorithms. A broad reinforcement learning (BRL) approach for IoT applications was proposed in

This research is funded by Guangdong Basic and Applied Research Foundation (Grant no.: 2021B1515130003), Science and Technology Development Fund of Macau (Grant nos.: 0091/2023/AMJ and 0026/2022/A) and the University of Macau (Grant no.: MYRG-GRG2023-00235-FST-UMDF). *(Corresponding author: Pak Kin Wong.)*

Shiorn Thalagala, Pak Kin Wong, Xiaozheng Wang, and Tianang Sun are with the Department of Electromechanical Engineering, University of Macau, Macau, China (e-mail: mc05396@connect.um.edu.mo; fstpkw@um.edu.mo; yc07429@connect.um.edu.mo; mc25058@connect.um.edu.mo).

All the source codes related to this study are available online at: https://github.com/ShironT/bcda.

[14], by substituting the critic network in basic Q-learning [15] with a BLS network. Although basic Q-learning is effective for RL problems characterized by low-dimensional discrete state-action spaces, its applicability is limited in scenarios involving high-dimensional continuous spaces [11]. However, the proposed BCDA framework can be utilized in control tasks involving high-dimensional continuous spaces. A BLS-based Q-learning for robotic control was investigated in [16], by employing offline training with pre-collected datasets. However, the requirement for available training data for weight estimation using ridge regression was highlighted as a challenge associated with BLS in RL. This presents a challenge for online training where pre-existing datasets are unavailable. The proposed BCDA framework is capable of dealing with online training without a pre-collected dataset. The work in [17] explores BLS for RL signal processing but does not address actor-critic architectures. In contrast, the BCDA framework pioneers the use of BLS in the critic network alongside a DNN-based actor to enhance the learning efficiency of actor-critic architecture. On the other hand, several other researchers attempted to enhance the efficiency and performance of DRL with different approaches instead of utilizing BLS. In [18], the authors proposed a transfer RL framework (RL-KS) that leverages prior knowledge from learned tasks to accelerate the learning of new tasks. Unlike RL-KS, which requires pre-trained models or historical data, BCDA operates in online settings without dependency on prior tasks, making it more adaptable to real-time control scenarios. Another study [19] explored a stochastic actor-critic framework that integrates exploration and exploitation strategies. Unlike deterministic critic updates via ridge regression used in the BCDA framework, this method employs stochastic optimization for both actor and critic networks. Although higher accuracies can be obtained due to the stochastic nature using the given method, computational efficiency could be reduced during continuous control tasks.

To our best knowledge, no prior research has been undertaken to explore the integration of BLS-based networks within actor-critic RL algorithms. Additionally, no prior study has been conducted on the exploration of applying BLS-based RL algorithms to continuous control tasks. In summary, the novelties of this study are as follows:

1) A novel actor-critic framework using a hybrid approach is proposed where both DNN and BLS are incorporated. This offers a unique architectural variation when compared to DRL and BRL.
2) The weight estimation approach in the proposed framework uses both gradient descent and ridge regression. The weights of the actor network are optimized through gradient descent while in the critic network, weights are estimated using ridge regression.
3) Several techniques are incorporated with the proposed framework to improve robustness and mitigate overfitting, including tuning the regularization strength, implementing an ensemble of BLS-based critic networks, and implementing dropout in the BLS-based critic network.
4) The proposed framework is integrated into the widely recognized DDPG framework and evaluated on seven simulated classic continuous control problems. Its performance is compared with that of the standard DDPG algorithm. Furthermore, the proposed framework is integrated with the TD3 and SAC algorithms and evaluated on two advanced simulated control problems to assess the efficiency enhancements relative to the original version of each algorithm.

The remainder of this brief paper is organized as follows. Section II details the construction and implementation of the proposed framework. In Section III, the proposed framework is evaluated. Finally, the conclusion is provided in Section IV.

## II. Broad Critic Deep Actor (BCDA) Framework

In this section, the construction of the BCDA framework built upon the DDPG algorithm (referred to “BCDA-DDPG algorithm”) is first described with details of the architectural components essential to its operation. Next, its implementation and functionality are discussed. Then, the discussion is extended to cover the application of the BCDA framework to the TD3 and SAC algorithms which results in BCDA-TD3 and BCDA-SAC, respectively. It should be noted that the architectural implementation of the BCDA framework is consistent across all three algorithms, as they all belong to the actor-critic class. Nevertheless, the only difference lies in the network parameter update component of each algorithm.

### *A. Framework of Proposed BCDA-DDPG Algorithm*

When compared to the standard DDPG algorithm, there are two significant differences in the proposed BCDA-DDPG algorithm. Firstly, the critic network is composed of a BLS rather than a DNN while the actor network is a DNN. Secondly, the actor network parameters are optimized using the gradient descent method while the critic network parameters are calculated using ridge regression.

The principal components of this framework are outlined as follows:

1) *Environment:* The environment represents the external system with which the agent interacts.
2) *Training buffer:* The training buffer is viewed as a short-term memory serving three main purposes. Firstly, it operates as the replay buffer, similar to its usage in the DDPG algorithm, where it stores the experiences derived from the interactions of the agent with the environment. Secondly, it stores the calculated targets for BLS training. Thirdly, it facilitates the computation of gradients of the Q-values with respect to the actor network parameters that are used for the optimization of the actor network. The training buffer records a sequence $R = [\{s_t, a_t, r_t, s_{t+1}, d\}, \{y_t\}]$ where $s_t, a_t$ and $r_t$ respectively represent the state, action, and reward at the time step $t$, $s_{t+1}$ is the subsequent state produced by the environment given action $a_t$. $d$ indicates whether the current state is terminal. $y_t$ denotes the target Q-values used for BLS training.

3) *Broad Critic Network (BCN):* The BCN processes inputs $\{s_t, a_t\}$ to generate the current Q-value at time step $t$, denoted as $Q_t$. The training of this network utilizes $y_t$ as target data. The calculation of the weights for the output layer of the critic network is achieved through the application of ridge regression. The BCN is illustrated in Fig. 1.

4) *Deep Actor Network (DAN):* The DAN is constructed using a straightforward DNN where its primary purpose is to approximate the optimal policy function.

5) *Target BCN (t-BCN) & Target DAN (t-DAN):* Both the t-BCN and t-DAN are initialized with parameters identical to those of the BCN and DAN, respectively. These target networks are updated simultaneously with the main networks.

### *B. Implementation of BCDA-DDPG Algorithm*

According to the framework outlined in Section II A, the implementation is carried out according to the following steps.

*1) Initialization:* Initially, both the DAN and BCN, including t-DAN and t-BCN, as well as the training buffer are initialized. The parameters of the BCN, denoted as an $\emptyset$, are randomly initialized and comprise $\{W_n^m, W_f, \beta_f, W_h, \beta_h\}$, where $W_n^m$ represents the weight of the final output layer, $W_f$ and $\beta_f$ are the weights and bias for the feature mapping nodes, and $W_h$ and $\beta_h$ correspond to the weights and bias for the enhancement nodes, respectively.

*2) Preparing Training Sample for BCN Training:* Training the BCN requires target values which are essentially functioning as labeled data. The primary objective of the BCN is to produce Q-values (approximating the action-value functions) upon the input of $\{s_t, a_t\}$. The aim is to refine the BCN such that the Q-values generated progressively align with the target Q-values. Generally, this optimization process involves minimizing the mean squared bellman error (MSBE) [20], which is mathematically represented as:

$$\mathrm{MSBE}_{\emptyset} = \mathbb{E}\left[\left(Q_{\emptyset}(s_t, a_t) - \left(r_t + \gamma_t\, d \max_{a_{t+1}} Q_{\emptyset}(s_{t+1}, a_{t+1})\right)\right)^2\right] \tag{1}$$

Hence, the target Q-values, $y_t$ can be defined as:

$$\mathrm{y_t} = \begin{cases} r_t, & \text{if } \mathrm{s}_{t+1} \text{ reaches } s_e \\ \gamma_t \max\limits_{a_{t+1}} Q_{\emptyset}(s_{t+1}, a_{t+1}), & \text{otherwise.} \end{cases} \tag{2}$$

where $\gamma_t$ is the discount factor in the range [0, 1], $s_e$ is terminal state and $\max\limits_{a_{t+1}} Q_{\emptyset}(s_{t+1}, a_{t+1})$ is computed using the t-BCN which is initialized with random weights, while $a_{t+1}$ is generated using the t-DAN.

In BCN training, $y_t$ serves as the target data, while $\{s_t, a_t\}$ sampled from the training buffer are the input data for training. Consequently, the BCN adjusts its weights to align with $y_t$, employing regression to approximate this target rather than directly minimizing the MSBE. This approach is different from the conventional weight optimization methodologies in typical RL algorithms that involve gradient descent to minimize MSBE.

*3) BCN Training:* The training input data, $\mathrm{X} = \{s_t, a_t\}$, and the target data, $\mathrm{Y} = \{\mathrm{y_t}\}$ are fed into the BCN network. First, feature extraction of X is performed to generate the i^th^ feature node, $Z_i$ as:

$$Z_i = \vartheta(XW_{f_i} + \beta_{f_i}), \quad i = 1, \dots, n. \tag{3}$$

where $W_{f_i}$ and $\beta_{f_i}$ are randomly generated weights and biases of the feature mapping nodes, respectively. $\vartheta(\cdot)$ is a nonlinear mapping function. Thus, all the feature nodes can be denoted as:

$$Z^n \equiv [Z_1, \dots, Z_n]. \tag{4}$$

Then, feature nodes are used to obtain enhancement nodes, $Z^n$ as:

$$H_j = \xi(Z^n W_{e_j} + \beta_{e_j}), \quad j = 1, \dots, m. \tag{5}$$

where $W_{e_j}$ and $\beta_{e_j}$ are randomly generated. $\xi(\cdot)$ is also a nonlinear mapping function. Thus, all the enhancement nodes can be denoted as:

$$H^m \equiv [H_1, \dots, H_m]. \tag{6}$$

Finally, all the feature nodes, $Z^n$ and enhancement nodes, $H^m$ can be concatenated for feeding into the final output layer as:

$$Y = [Z^n | H^m] W_n^m. \tag{7}$$

Take $A_n^m = [Z^n | H^m]$, then $Y = A_n^m W_n^m$.

$Y$ is the output of the BCN while $W_n^m$ is the weight of the final output layer. Here, ridge regression approximation is utilized to find the weight $W_n^m$. It can be calculated using the pseudoinverse of the $A_n^m$ as follows:

$$W_n^m = (A_n^m)^+ \tag{8}$$

$$W_n^m = (A_n^m\, (A_n^m)^T + \lambda \boldsymbol{I})^{-1}\, (A_n^m)^T\, Y \tag{9}$$

If the above method does not provide sufficient accuracy during the training of the BCN, additional enhancement nodes can be added to increase the accuracy.

Let's assume that one additional enhancement node $H_{m+1}$ is added to the network. Then using (5):

$$H_{m+1} = \xi(Z^n W_{e_{m+1}} + \beta_{e_{m+1}}) \tag{10}$$

where $W_{e_{m+1}}$ and $\beta_{e_{m+1}}$ are the weight and the bias of the additional enhancement node. According to theory in [21] pseudoinverse of the new matrix is calculated as:

$$A_n^{m+1} = \begin{bmatrix} (A_n^m)^+ - \boldsymbol{D}\,\boldsymbol{B}^{\boldsymbol{T}} \\ \boldsymbol{B}^{\boldsymbol{T}} \end{bmatrix} \tag{11}$$

where $\boldsymbol{D} = (A_n^m)^+ H_{m+1}$, and

$$\boldsymbol{B}^{\boldsymbol{T}} = \begin{cases} (\boldsymbol{C})^+, & if\ \boldsymbol{C} \neq 0 \\ (1 + \boldsymbol{D}^{\boldsymbol{T}}\,\boldsymbol{D})^{-1}\,\boldsymbol{D}^{\boldsymbol{T}}\,(A_n^m)^+, & if\ \boldsymbol{C} = 0 \end{cases} \tag{12}$$

where $\boldsymbol{C} = H_{m+1} - (A_n^m)\,\boldsymbol{D}$.

Hence, the weight of the final output layer after adding new enhancement node is given by:

$$W_n^{m+1} = \begin{bmatrix} (W_n^m) - \boldsymbol{D}\,\boldsymbol{B}^{\boldsymbol{T}}\,\boldsymbol{Y} \\ \boldsymbol{B}^{\boldsymbol{T}}\,\boldsymbol{Y} \end{bmatrix}. \tag{13}$$

Computation of the pseudoinverse of the additional enhancement nodes is enough instead of computing the entire $A_n^{m+1}$ matrix. Consequently, if adding enhancement nodes does not yield the anticipated improvement in the accuracy, additional feature nodes can be added similarly to further increase the accuracy [9].

*4) Updating DAN:* Within the BCDA algorithm, the actor network is updated similarly to the approach utilized in the DDPG algorithm as detailed in [11].

*5) Updating t-BCN & t-DAN:* The t-BCN and t-DAN are updated as follows:

$$\emptyset_{t-BCN} \leftarrow \rho\emptyset_{t-BCN} + (1-\rho)\emptyset_{BCN} \tag{14}$$

$$\emptyset_{t-DAN} \leftarrow \rho\emptyset_{t-DAN} + (1-\rho)\emptyset_{DAN} \tag{15}$$

It should be noted that different strategies can be utilized for the weight optimization in the BCN in addition to the ridge regression. The proposed framework of the BCDA-DDPG algorithm is depicted in Algorithm 1.

### *C. BCDA-TD3 and BCDA-SAC Algorithms*

To implement the BCDA framework into the TD3 and SAC algorithms, the critic networks in both algorithms are replaced with BLS networks while retaining DNNs for the actor. For TD3, which employs twin critics to mitigate overestimation bias, the twin DNN critics are substituted with twin BCNs. These BCNs take the state-action pair as input and estimate Q-values using ridge regression. It enables rapid updates without gradient descent. During training, target Q-values $y_t$ are computed using the t-DAN and two t-BCNs as follows:

$$y_t = r_t + \gamma_t \cdot \min_{i=1,2} Q_{\text{t-BCN}}^{(i)}\big(s_{t+1}, \text{t-DAN}(s_{t+1})\big). \tag{16}$$

This approach of taking the minimum Q-value between the twin BCNs ensures stability. The actor network remains a DNN and is updated via gradient ascent to maximize the Q-values estimated by the first BCN. Target networks are softly updated using (14) and (15) to preserve delayed update mechanism of the original TD3 algorithm. This hybrid architecture reduces computational cost significantly, as ridge regression in BLS eliminates the iterative backpropagation required for DNN-based critics. Also, the concept of twin critics in the original TD3 algorithm retains the robustness against overestimation errors.

In the proposed BCDA-SAC, the traditional DNN-based critics are replaced with two BCNs to retain the robustness of SAC algorithm against overestimation in Q-value. These BCNs independently estimate Q-values, and the minimum of their outputs is used for policy updates and target calculations, aligning with twin-critic design of SAC. The stochastic actor, which outputs parameters for a Gaussian policy, remains a DNN to enable gradient-based optimization of the policy mean and variance. During critic updates, target Q-values $y_t$ incorporate entropy regularization to balance reward maximization and exploration as follows:

$$y_t = r_t + \gamma_t \cdot (Q - \alpha \log \pi_{t-\text{DAN}}(s_{t+1})) \tag{17}$$

$$Q = \min_{i=1,2} Q_{\text{t-BCN}}^{(i)}(s_{t+1}, a_{t+1}) \tag{18}$$

where $a_{t+1} \sim \pi_{\text{t-DAN}}(s_{t+1})$, $\alpha$ denotes entropy coefficient, and $\log \pi_{t-\text{DAN}}$ denotes the logarithm of the probability density of the action $a_{t+1}$. The BCNs are trained via ridge regression to approximate entropy-augmented Q-values. It accelerates convergence compared to gradient-based DNN critics. The actor is optimized to maximize a combination of Q-values and policy entropy, with the $\alpha$ dynamically adjusted to maintain a target entropy level. Target networks are used only for the BCN critics and are updated using (14) and (15). By integrating BLS into SAC, the algorithm retains its exploration advantages while reducing computational costs.

### *D. Rationale for Choosing BLS for Critic Networks Rather than Actor Networks*

The BLS is uniquely suitable for critic networks due to its strength in regression tasks through ridge regression. It aligns with the role of the critic network in actor-critic algorithms, which is the estimation of Q-values. In contrast, the actor network learns a policy function (i.e., a probability distribution or deterministic action), which requires gradient-based optimization to maximize expected returns.

Two key factors justify this design. The first is the optimization compatibility. The Q-value estimation of the critic network is a regression problem where BLS excels due to its closed-form solution. This avoids iterative gradient optimization and reduces training time. The policy optimization of the actor network relies on gradient ascent through the policy gradient theorem [10], which inherently requires differentiable parameter updates. With its fixed random feature mappings and non-differentiable pseudoinverse operation, BLS cannot directly propagate gradients to refine policy outputs. The second factor is the balance between exploration and stability. Actor networks benefit from stochastic exploration (e.g., Gaussian noise in DDPG [11]), which is challenging to implement with BLS due to its deterministic, non-probabilistic outputs. BLS-based actor networks may require additional probabilistic layers (e.g., Gaussian output with learnable variance), which introduce complexity that could diminish the efficiency advantages of BLS. Thus, the hybrid architecture (i.e., BLS critic with DNN actor) balances the computational efficiency (via BLS regression) and gradient-based policy optimization (via DNNs). Future work may explore BLS-based actors for stochastic policies without compromising efficiency, but it remains an open problem.

## III. Experiments and Discussions

In this section, the numerical evaluation of the proposed approach is outlined, followed by an analysis of the obtained results.

### A. Evaluation Methodology

Initially, to evaluate the generalization, adaptability, and efficiency of the proposed BCDA-DDPG algorithm, seven MuJoCo [22] continuous control tasks (Fig. 2) integrated with OpenAI Gym [23] are utilized:

1) *Inverted pendulum (INP):* A classic control task where the agent must balance a pendulum on a cart.
2) *Inverted double pendulum (DIN):* A more challenging variant of INP that requires stabilization of two linked pendulums.
3) *Reacher (REA):* A 2-DOF robotic arm tasked with reaching a randomly positioned target.
4) *Pusher (PUS):* A robotic arm that must push an object to a target location while avoiding obstacles.
5) *Ant (ANT):* A quadruped robot tasked with locomotion in a planar environment.
6) *Half cheetah (HCH):* A high-dimensional bipedal robot aiming to achieve high-speed forward motion.
7) *Humanoid (HUM):* A high-dimensional and complicated humanoid robot tasked with stable walking.

These tasks include a wide range of state-action space dimensionalities, reward structures, and environmental dynamics. For instance, the HUM task involves 376-dimensional state spaces and 17-dimensional action spaces, while INP operates in a simpler 4-dimensional state space. This diversity ensures a comprehensive assessment of the scalability and adaptability of the BCDA framework.

To evaluate BCDA-TD3 and BCDA-SAC, two of the most advanced tasks (i.e., HCH and HUM) are selected. Prior studies [12], [13] validated the superiority of the original SAC and TD3 over DDPG in complex tasks. This study aims to demonstrate that augmenting these algorithms with the BCDA framework enhances learning efficiency without compromising accuracy.

The experimental evaluations are performed using an NVIDIA Tesla T4 GPU and all algorithms are implemented in Python using TensorFlow [24]. Each experiment is conducted across five distinct random trials, extending through 100,000 timesteps. Performance metrics are assessed at intervals of 500 timesteps, during which the algorithms are executed without including action noise, and the average reward is recorded. Additionally, the average training duration is also recorded.

The base BCN (b-BCN) has 10 feature nodes and 500 enhancement nodes. Across all the variants of BCNs, shrinkage coefficient, regularization factor, and learning rates are consistently set as 0.8, $2^{-30}$, and 0.005 respectively. All the DNNs in both BDCA and DDPG algorithms are three-layer feed-forward networks with 256 hidden nodes in each layer and rectified linear units (ReLU) [25] as the activation function between each layer.

The incremental learning (IL) feature within the BLS is significantly beneficial in enhancing the performance of the training process [9]. To evaluate this, four training schemes are designed to train the REA using the BCDA algorithm: (a) Using the b-BCN without IL (referred to as “IL Scheme-1”), (b) IL by adding 5 feature nodes and 100 enhancements nodes to the b-BCN (referred to as “IL Scheme-2”), (c) IL by adding 5 feature nodes and 300 enhancements nodes to the b-BCN (referred to as “IL Scheme-3”), and (d) IL by adding 10 feature nodes and 300 enhancement nodes to the b-BCN (referred to as “IL Scheme-4”).

For IL in different control tasks, a general parameter selection framework can be defined as follows. In low-dimensional tasks (e.g., INP), prioritizing fewer feature nodes (≤10) and moderate enhancement nodes (≤200) is recommended to balance model complexity and efficiency. Conversely, for high-dimensional tasks (e.g., HUM), expanding the feature nodes (10–20) and enhancement nodes (200–500) enhances the representational capacity of the critic network. During incremental tuning, training should begin with a base configuration, such as 10 feature nodes and 500 enhancement nodes. The validation loss must then be monitored throughout training, with additional nodes incrementally added if performance is reduced.

### B. Learning Accuracy and Efficiency

Fig. 3(a)-(g) illustrates the learning curves for the INP, DIN, REA, PUS, ANT, HCH, and HUM control tasks, respectively across the five trials, indicating the moving average of the average reward accumulated throughout the learning trajectories for each task. A window size of 10 is utilized to compute the moving averages. The shaded regions around the lines represent the half-standard deviation of the average evaluations across the five trials. It provides an insight into the variability of the performance across 5 trials.

In all control tasks, the BCDA-DDPG algorithm exhibits a significantly steeper learning curve during the initial timesteps compared to the DDPG algorithm, indicating a faster convergence towards the optimal reward. However, since the original DDPG algorithm struggles to converge to the optimal reward value (within $10^5$ timesteps) in complex control tasks such as ANT, HCH, and HUM, the BCDA-DDPG algorithm also does not reach the optimal reward value. Table I shows the maximum average reward over 5 trials obtained using BCDA-DDPG and DDPG algorithms on all control tasks. It can be observed that BCDA-DDPG offers a higher maximum average reward on several control tasks. This suggests that the BCDA framework is successful in learning the continuous control tasks and offers a faster learning rate.

Table II presents the average training duration across five trials for both DDPG and BCDA-DDPG agents. BCDA-based algorithm demonstrates a significant reduction in training time compared to the original DDPG algorithm due to the rapid learning capabilities of the BLS. Thus, it is demonstrated that integrating BLS with DNN under the actor-critic framework can enhance efficiency without compromising learning accuracy.

Fig. 4 illustrates the learning curves of the HCH and HUM tasks, comparing the performance of BCDA-TD3 with the original TD3 algorithm. The results demonstrate that BCDA-TD3 achieves higher accuracy and faster learning rate in both control tasks. Additionally, BCDA-TD3 exhibits reduced fluctuations in reward variance compared to the original TD3 which highlights improved stability. Similarly, Fig. 5 presents the learning curves for the same tasks using BCDA-SAC and the original SAC algorithm. These results validate that the BCDA framework improves both efficiency and accuracy even under advanced actor-critic algorithms like SAC and TD3. Future studies will further explore the trade-off among speed, stability, and performance in complex tasks such as the boiler-turbine tracking control discussed in [26].

### C. Evaluation of Robustness of BCDA Framework

To maintain conciseness and avoid redundancy of the paper, only the INP task is used for robustness evaluation. The learning curves in Fig. 3 are generated by training the algorithm with the robustness enhancement techniques discussed in this section.

Ridge regression used in BLS inherently incorporates regularization to mitigate overfitting, but the regularization strength (λ) requires further optimization. To investigate this, an additional experiment is conducted using the INP task with varying regularization factors ($\lambda = [2^{-20}, 2^{-30}, 2^{-40}]$). The results in Fig. 6 demonstrate that stronger regularization ($\lambda = 2^{-20}$) reduces post-convergence reward variance by 18.12% compared to the baseline ($\lambda = 2^{-30}$), even there is a slight compromise in learning speed (Table III). Conversely, weaker regularization ($\lambda = 2^{-40}$) induces instability. Nevertheless, the scenario with $\lambda = 2^{-20}$ still displays moderate fluctuations after convergence. Since further increasing the regularization factor can slow learning, $\lambda = 2^{-20}$ is chosen to explore alternative methods of enhancing robustness.

Two additional robustness enhancement methods are investigated. The first is the ensemble BCNs. Inspired by Ref.[27], an ensemble of 5 architecturally similar BCNs are deployed with randomized feature and enhancement node initialization. The final Q-value is computed as the median of ensemble outputs. This approach suppresses outlier Q-estimates, improving the stability of INP. Next, dropout [28] (rate = 0.2) is integrated into the enhancement nodes during BCN training. Dropout mitigates co-adaptation of enhancement nodes and acts as implicit regularization. Combined with $\lambda = 2^{-20}$, these further enhancements achieve the lowest variance in long-term INP trials. Fig. 7 compares $\lambda = 2^{-20}$ versus $\lambda = 2^{-20}$ with ensemble BCN and dropout. Fig 7(b) shows a clear reduction of variation in the average reward after the point of convergence. These results show the efficacy of advanced regularization in BCDA framework. Similarly, the same method can be repeated on the other control tasks.

### D. Analysis of Incremental Learning Evaluation Results

Fig. 8 illustrates the impact of IL on the REA task. Scheme-3 (5 feature nodes + 300 enhancement nodes) achieves the highest average reward due to an optimal balance between feature diversity and model complexity. The 5 added feature nodes extract spatial representations of the state-action space and enhance the capacity of BCNs to distinguish subtle differences in Q-values. The 300 enhancement nodes introduce nonlinear transformations of these features and enable precise approximation of Q-values without overfitting.

In contrast, in Scheme-1 (No IL), the model capacity is limited, resulting in underfitting. It fails to capture high-precision Q-value relationships. In Scheme-2, insufficient enhancement nodes restrict nonlinear expressiveness. It leads to slower convergence. Moreover, in Scheme-4, excessive feature nodes cause redundant representations. It increases variance and destabilizes training, which is evident in erratic reward fluctuations.

## IV. CONCLUSION

Improving efficiency is crucial in online RL for continuous control problems. This brief study introduces an innovative variation to the actor-critic RL architecture by incorporating BLS with DNN in a hybrid manner. Within this architecture, the critic network utilizes BLS while the actor network is implemented using DNN. Furthermore, ridge regression is employed to determine the parameters of the critic network, while the actor network optimizes weights using gradient optimization.

Although BLS was applied in RL algorithms in previous studies, its application within actor-critic algorithms, especially for learning continuous control problems, has not been explored. The results reveal that the BCDA framework-based control approach not only enhances modeling and decision-making capabilities but also significantly reduces computational complexity compared to the three widely recognized actor-critic RL algorithms (i.e., DDPG, SAC, and TD3).

This study opens up multiple avenues for future research. Although BCDA improves computational efficiency over DDPG, SAC, and TD3 algorithms, its scalability to large-scale datasets (e.g., industrial systems generating terabytes of sensor data) require further optimization. Techniques, such as sparse BLS node pruning [29], distributed ridge regression via MapReduce [30], and quantization-aware training [31], could be considered to reduce memory and computation costs. Additionally, integrating BCDA with edge computing frameworks would enable deployment on resource-constrained hardware. These innovations will be essential for the application of the BCDA framework to real-time applications. Ultimately, this study suggests that incorporating the proposed hybrid BCDA framework with actor-critic algorithms could further improve the efficiency of RL algorithms.

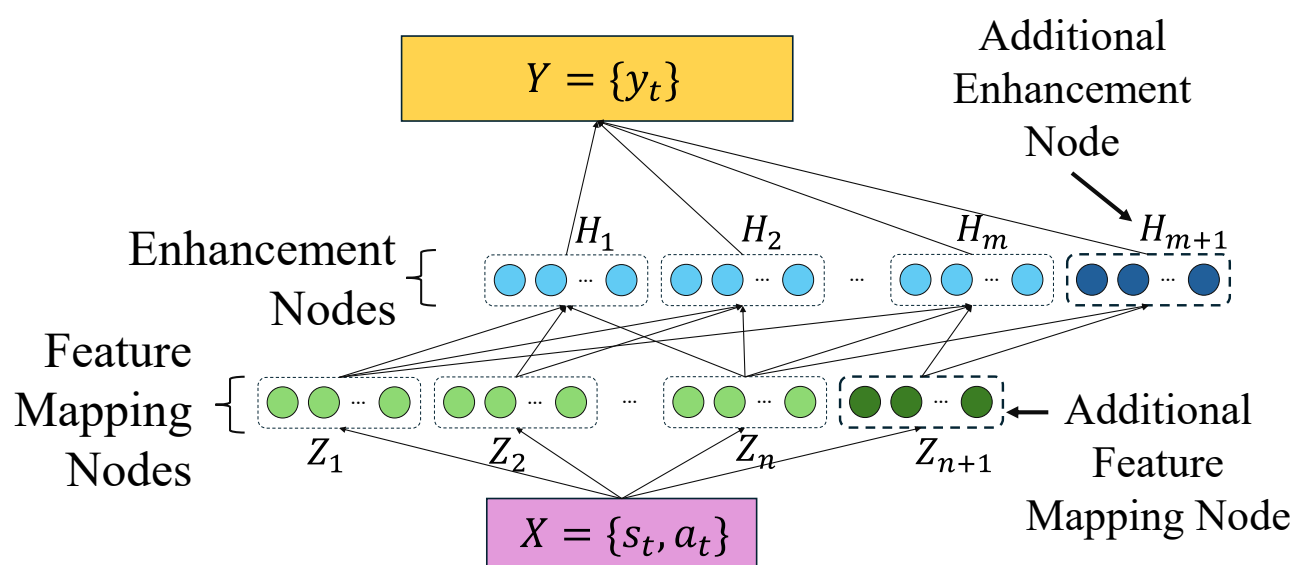

**Fig. 1.** Schematic of BCN architecture.

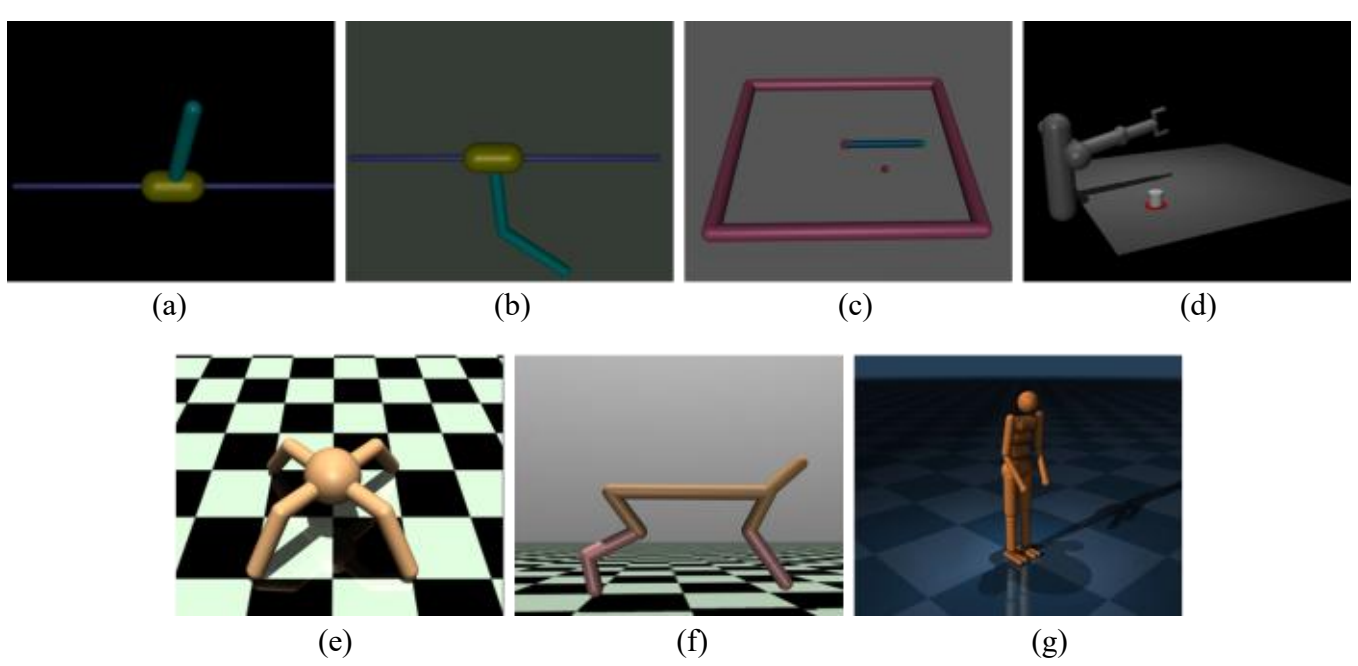

**Fig. 2.** Control tasks used for evaluations: (a) INP, (b) DINV, (c) REA, (d) PUS, (e) ANT, (f) HCH, and (g) HUM.

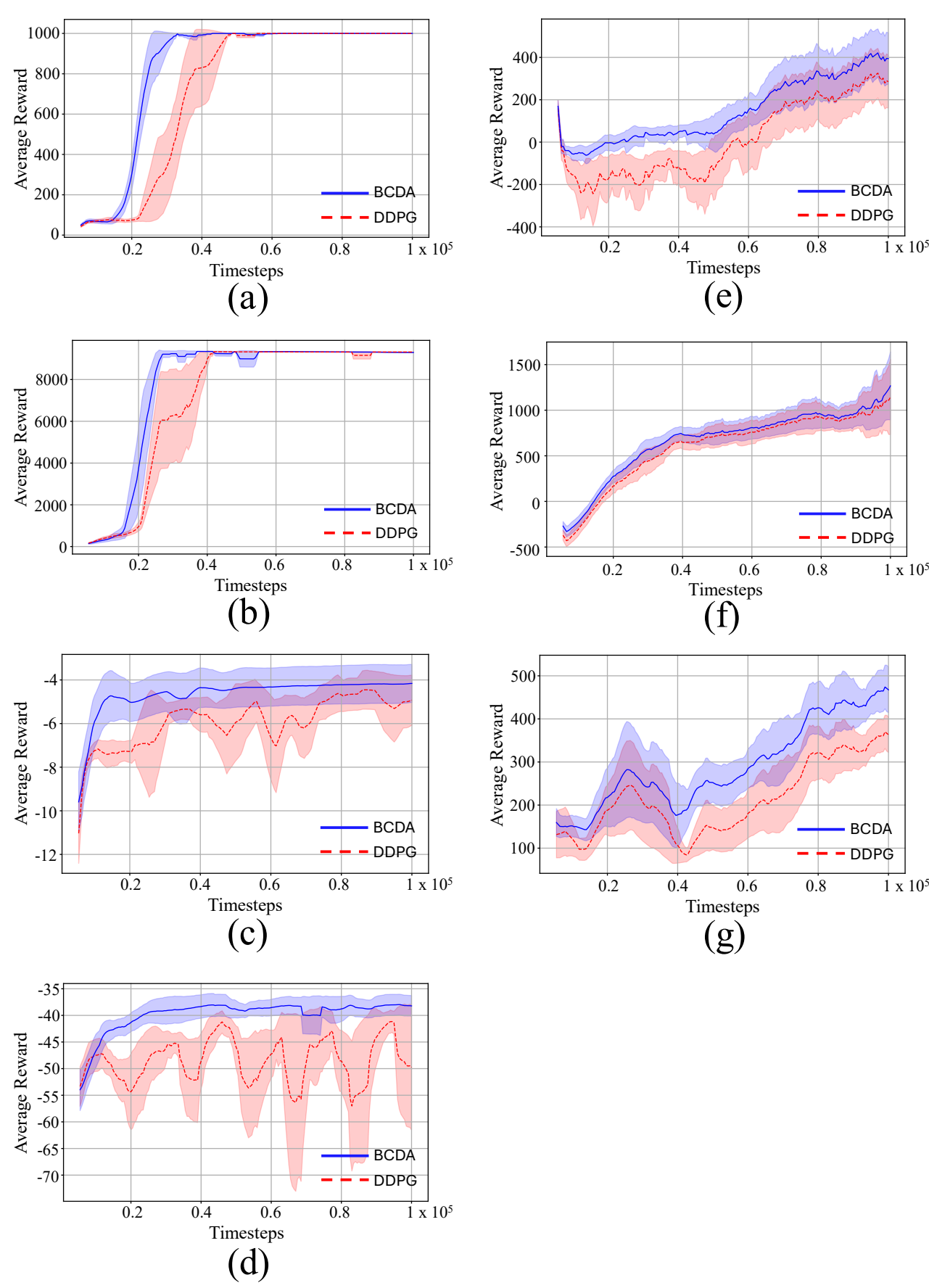

**Fig. 3.** Learning curves of (a) INP, (b) DINV, (c) REA, (d) PUS, (e) ANT, (f) HCH, and (g) HUM across 5 trials of $10^5$ timesteps. The shaded region in each line represents half of the standard deviation across the 5 trials.

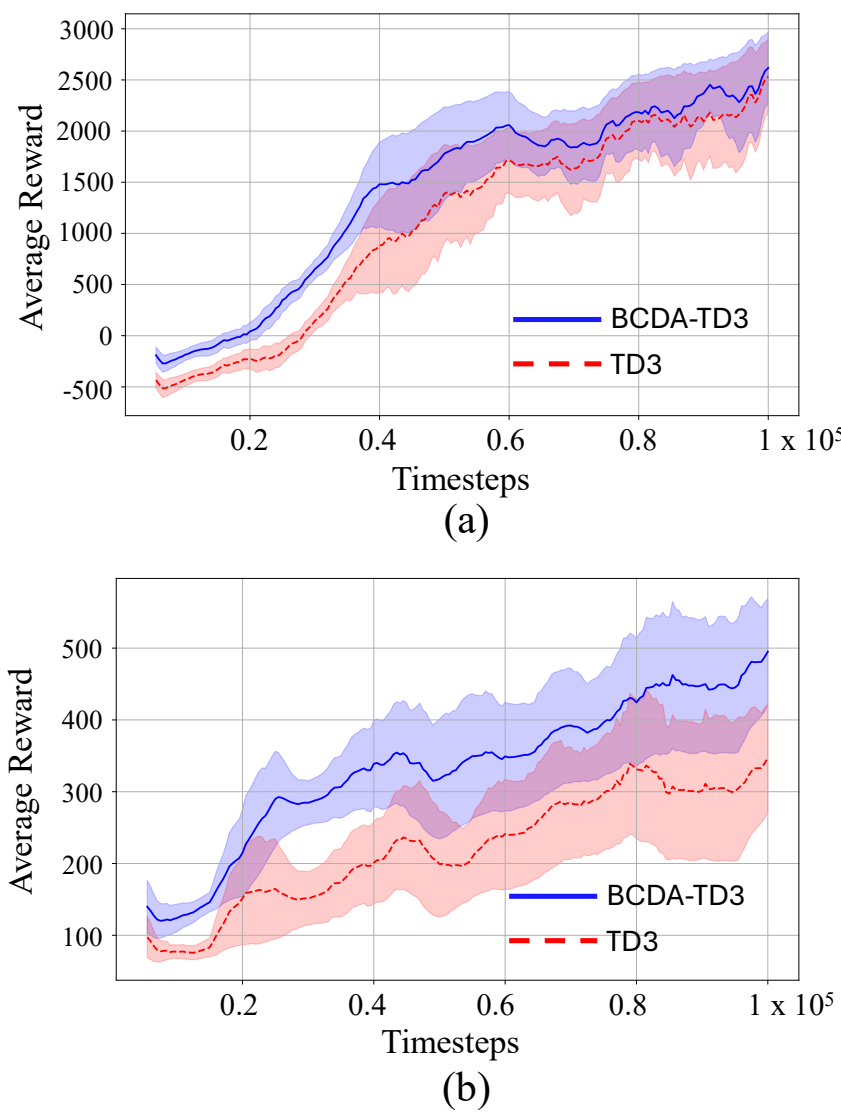

**Fig. 4.** Learning curves of (a) HCH and (b) HUM across 5 trials of $10^5$ timesteps with BCD-TD3 and TD3. The shaded region in each line represents half of the standard deviation across the 5 trials.

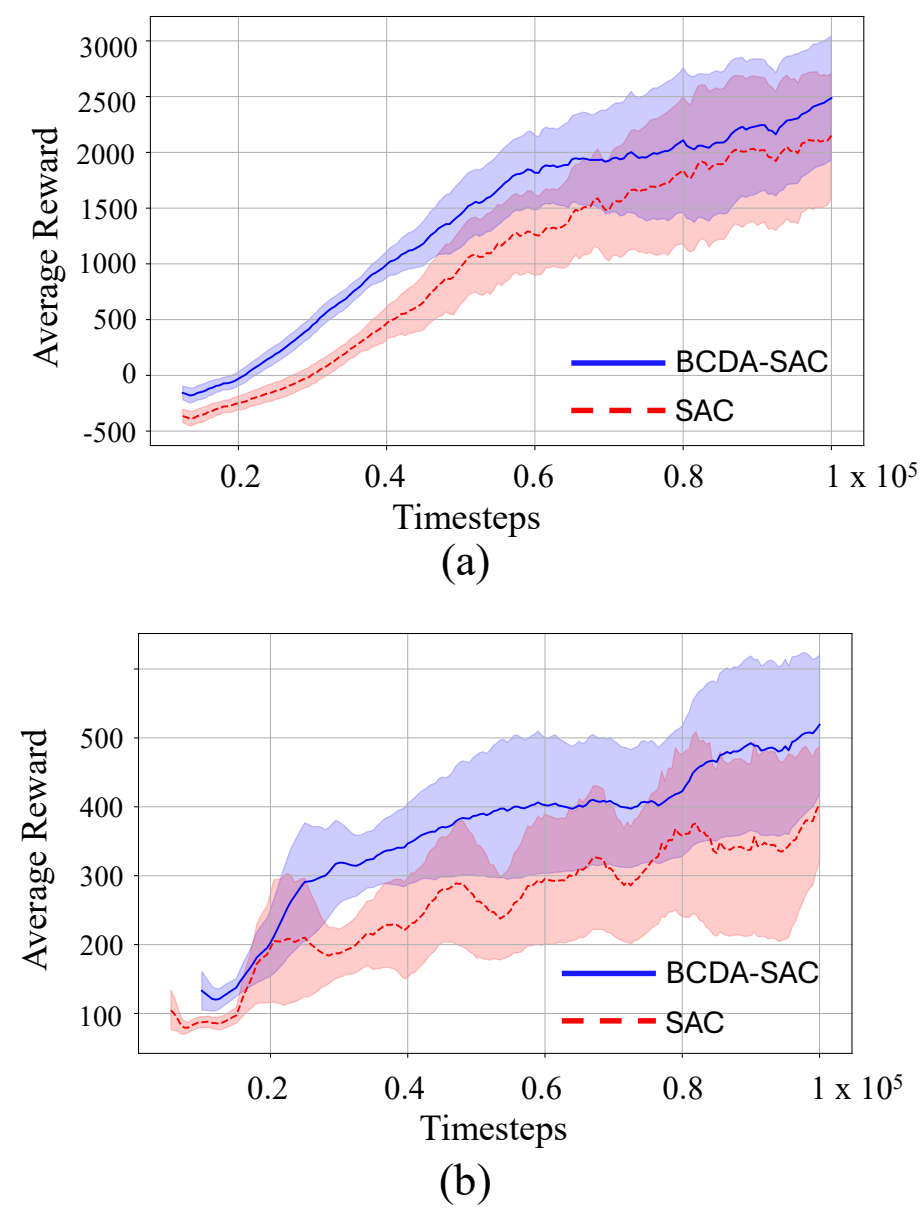

**Fig. 5.** Learning curves of (a) HCH and (b) HUM across 5 trials of $10^5$ timesteps with BCD-SAC and SAC. The shaded region in each line represents half of the standard deviation across the 5 trials.

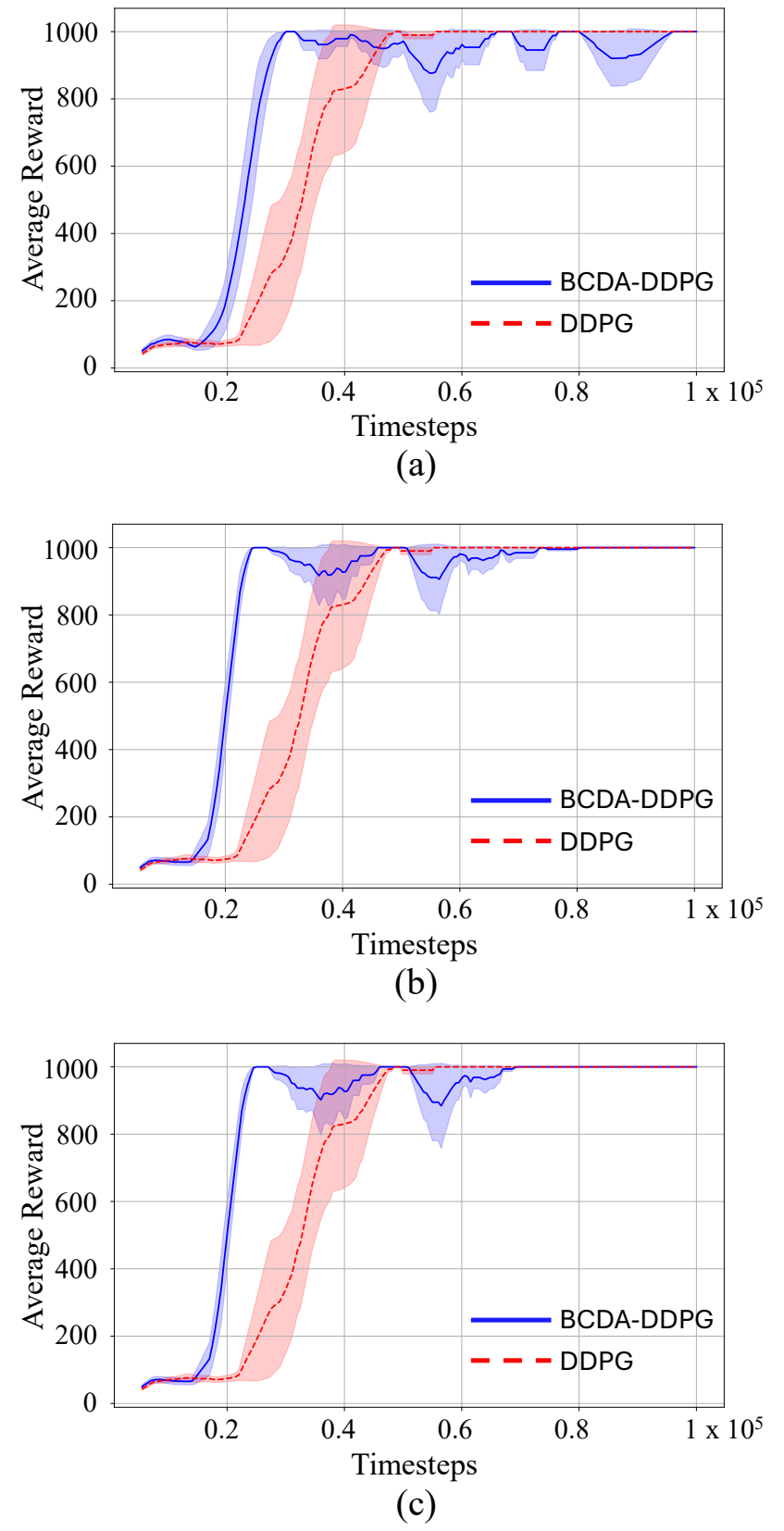

**Fig. 6.** Learning curve by varying $\lambda$ in INV control task with the BCDA-DDPG algorithm. (a) $\lambda = 2^{-40}$, (b) $\lambda = 2^{-30}$, and (c) $\lambda = 2^{-20}$.

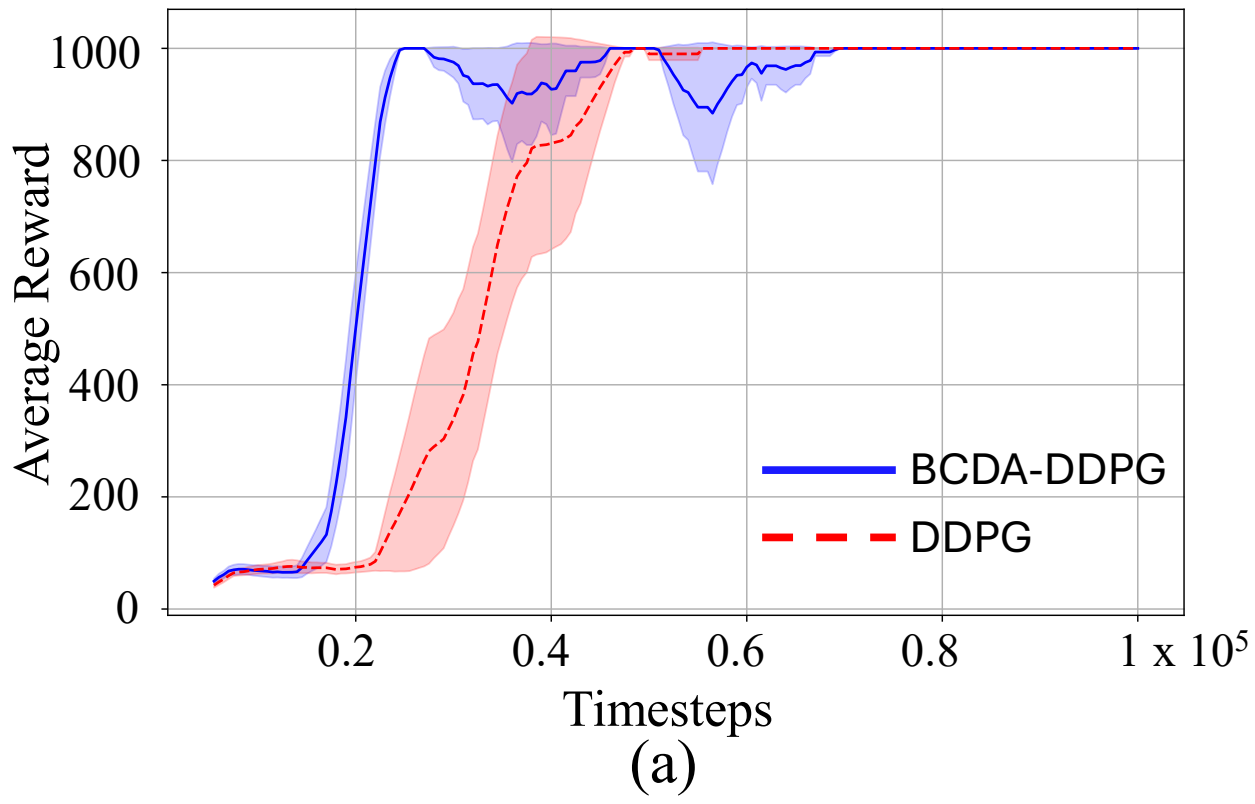

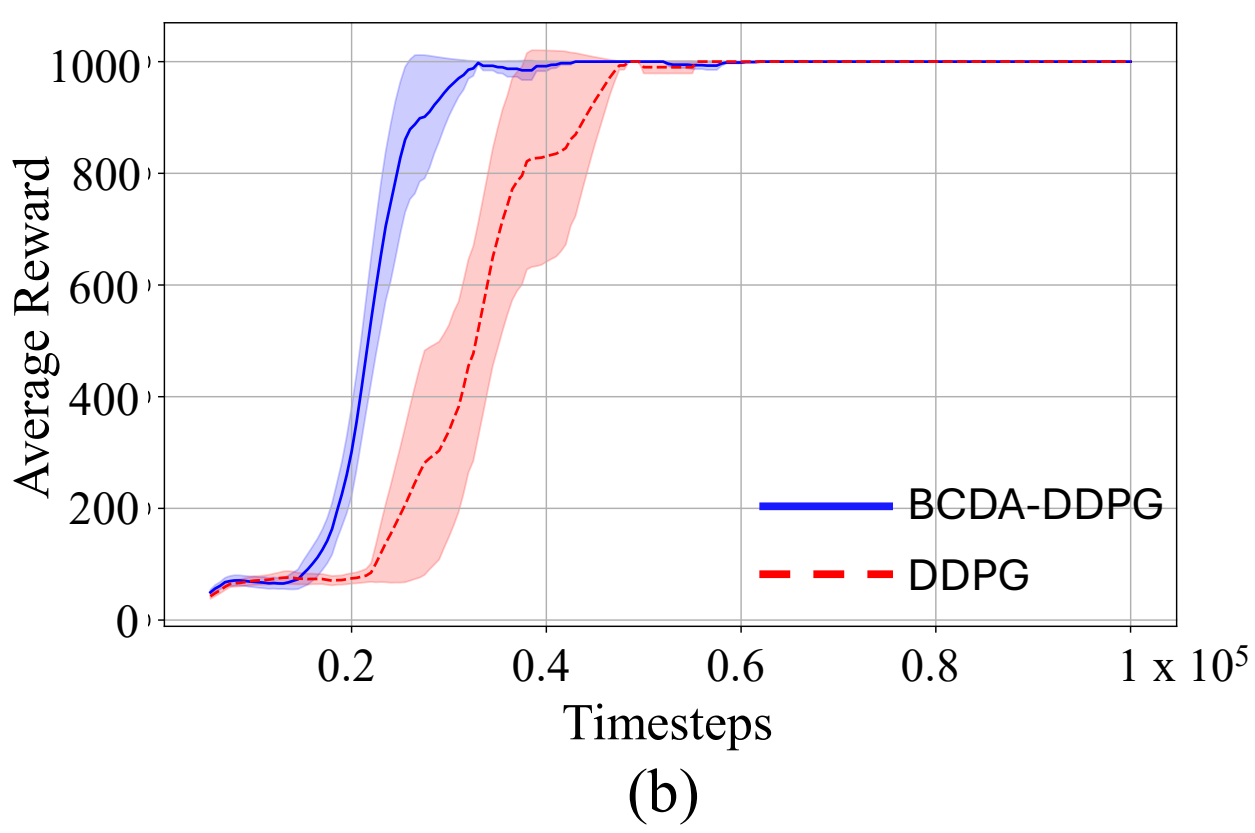

**Fig. 7.** Learning curve of INV control task with BCDA-DDPG algorithm with (a) $\lambda = 2^{-20}$ and (b) $\lambda = 2^{-20}$ together with ensemble BCN and dropout.

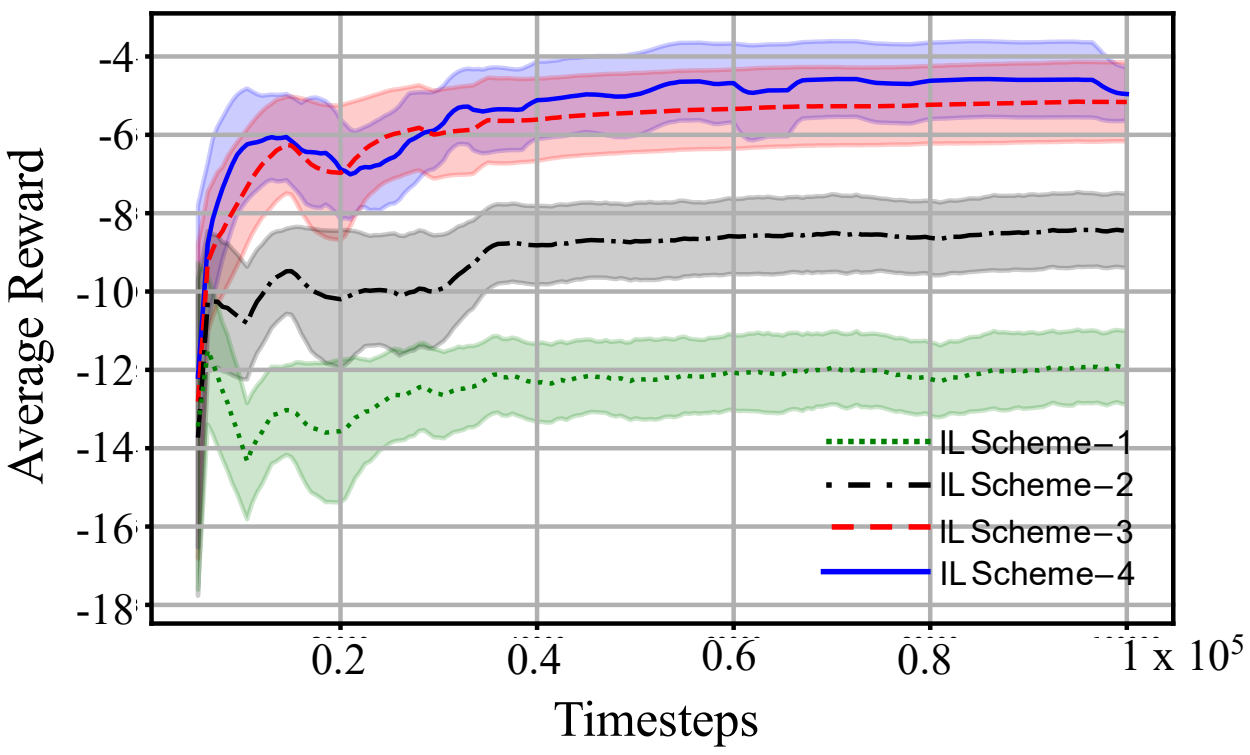

**Fig. 8.** Learning curves of REA when trained using BCDA-DDPG algorithm on different IL schemes.

---

**Algorithm 1** Broad Critic Deep Actor (BCDA).

---

1: Initialize BCN & DAN with random parameters
2: Initialize t-BCN & t-DAN
3: Initialize training buffer $\mathfrak{p}$ with batch size $B$
4: **for** episode ← 1 to total episodes **do**
5: Observe Initial state $s$
6: Initialize time step t = 0
7: **if** $t < T$ **or** $t = T$ **then**
8: Sample an action $a$ randomly
9: **else** $t > T$ **then**
10: Compute action $a$ using DAN
11: **end if**
12: Execute $a$ in the environment
13: Observe next state $s'$, reward $r$, and $d$
14: Store $(s, a, r, s', d)$ in the training buffer $\mathfrak{p}$
15: If $s'$ reaches terminal, reset environment state
16: **if** training buffer size > $B$ **and** $t > T$ **then**
17: Randomly sample $\{(s, a, r, s', d)\}$ from $\mathfrak{p}$
18: Calculate $y_t$ using (2)
19: Train the BCN with X and $Y$
20: **while** the BCN accuracy is unsatisfactory
21: Perform incremental learning
22: **end while**
23: Finish training BCN and obtain $\phi$
24: Compute gradients for $Q(s, a)$ with respect to $\theta$
25: Update policy network as in [11]
26: Update target networks using (14) & (15)
27: **end if**
28: **end for**

---

TABLE I
MAXIMUM AVERAGE REWARD OF ALL CONTROL TASKS IN 10000 TIMESTEPS OVER 5 TRIALS WITH BCDA-DDPG ALGORITHM

| Control Task | DDPG | BCDA-DDPG |
| --- | --- | --- |
| INP | 1000 | 1000 |
| DIN | 9359.95 | 9359.97 |
| REA | -1.39 | -1.36 |
| PUS | -31.27 | -27.43 |
| ANT | 995.51 | 995.51 |
| HCH | 3451.64 | 3504.00 |
| HUM | 928.105 | 928.10 |

TABLE II
AVERAGE TRAINING TIME ACROSS 5 TRIALS IN THE TWO CONTROLS TASKS WITH BCDA-DDPG ALGORITHM

| **Control Task** | **DDPG (s)** | **BCDA-DDPG (s)** | **Efficiency Improvement of BCDA-DDPG Relative to DDPG (%)** |
|---|---|---|---|
| INP | 1324.22 | 429.12 | 67.59 |
| DIN | 1309.80 | 511.84 | 60.92 |
| REA | 613.77 | 256.98 | 58.13 |
| PUS | 700.51 | 425.63 | 39.24 |
| ANT | 2433.38 | 1648.25 | 32.27 |
| HCH | 1519.29 | 985.32 | 35.15 |
| HUM | 885.37 | 689.25 | 22.08 |

TABLE III
QUANTITATIVE COMPARISON OF REWARD VARIANCE AND AVERAGE TRAINING TIME ACROSS $\lambda$

| $\boldsymbol{\lambda}$ | **Training time (s)** | **Variance of the reward among 5 seeds** | **Percentage difference of the variance compared to $\boldsymbol{\lambda = 2^{-30}}$** |
|---|---|---|---|
| $\boldsymbol{\lambda = 2^{-30}}$ | 485.52 | 12419.54 | N/A |
| $\boldsymbol{\lambda = 2^{-20}}$ | 521.15 | 6721.61 | 18.12 % decrease |
| $\boldsymbol{\lambda = 2^{-40}}$ | 435.25 | 19571.70 | 89.59 increase |